\documentstyle[11pt,psfig]{article}
\begin{document}
\title{\LARGE\bf A Learning Approach to Natural Language Understanding}
\author{\large\it Roberto Pieraccini \and \it Esther Levin}
\date{\large AT\&T Bell Laboratories \\
600 Mountain Avenue \\
Murray Hill, NJ 07974}
\maketitle
\begin{abstract}
In this paper we propose a learning paradigm for the problem
of understanding spoken language. The basis of the work is in a formalization
of the understanding problem as a communication problem. This results in
the definition of a stochastic model of the production of speech or
text starting from the meaning of a sentence. The resulting understanding algorithm 
consists in 
a  Viterbi maximization procedure, analogous to that commonly
used for recognizing speech. The algorithm was implemented for building
a  module, called {\em conceptual decoder} 
for the decoding of the conceptual content of sentences in
an airline information domain. The decoding module is the basis on
which a complete prototypical understanding system was implemented 
and whose performance are discussed in the paper. The problems, the possible
solutions and the future directions of the learning approach to
language understanding are also discussed in this paper.
\end{abstract}
\section{Introduction}
\begin{figure}[ht]
\centerline{\psfig{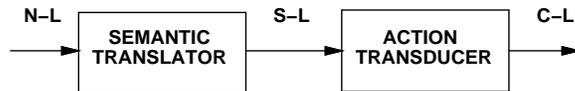}}
\caption{Understanding as a translation process \label{fig:transl}}
\end{figure}

A natural language understanding system is a machine
that produces an action as the result of an input sentence
(speech or text).
There are examples~\cite{gorin-91} of systems
that are able of modeling and learning the relationship between the
input sentence and the action in a direct way. 
However, when the task is rather complicated, i.e. the set of possible actions
is extremely large, we believe that it is necessary to rely on a 
intermediate symbolic representation. Fig.~\ref{fig:transl}
depicts a natural language understanding as composed of
two components. The first, called
{\it semantic translator} analyzes the input sentence in natural language
{\it (N-L)} and
generates a representation of its meaning in a formal semantic language
{\it (S-L)}.
The {\it action transducer} converts the meaning
representation into statements of
a given computer language {\it (C-L)} for executing the required action.\\

Although there are several and well established ways~\cite{allen-87} of performing the
semantic translation with relatively good performance,
we are interested in investigating the possibility
of building a machine that can learn how to do it from the
observation of examples.
Traditional {\em non-learning} methods are based on grammars (i.e. set of
rules) both at the syntactic and semantic level. Those grammars are
generally designed by hand. Often the grammar designers rely on corpora
of examples for devising the rules of a given application. But the
variety of expressions that are present in a language, even though it
is restricted to a very specific semantic domain, makes the task 
of refining a given set of rules an endless job. Any additional set of 
examples may lead to the introduction of new rules, and while the
rate of growth of the number of rules decreases with the number
of examples, larger and larger amounts of data must be analyzed for
increasing the coverage of a system.
Moreover,  if a new different application has to be designed,
very little of the work previously done can be generally exploited. 

The situation
is even more critical for spoken rather than written language. Written language
generally follows {\em standard} grammatical rules more strictly than spoken
language, that is often ungrammatical and idiomatic. Besides,
in spoken language, there are phenomena like false starts and
broken sentences that do not appear in written language. The 
following is a real example from a
corpus of
dialogues~\cite{madcow-92} within the airline information domain, 
(DARPA ATIS project~\cite{price-90}, see section~\ref{sect:implement}).
\begin{quote}
{\it FROM uh sss FROM THE PHILADELPHIA AIRPORT um AT ooh
THE AIRLINE IS UNITED AIRLINES AND IT IS FLIGHT NUMBER ONE NINETY
FOUR ONCE THAT ONE LANDS I NEED GROUND TRANSPORTATION TO uh
BROAD STREET IN PHILELD PHILADELPHIA WHAT CAN YOU ARRANGE FOR
THAT}~\footnote{this sentence was cited by Victor Zue, MIT, during
the 5th DARPA Workshop on Speech and Natural Language, Harryman, NY,
Dec. 1991.}
\end{quote}
It is clear from this example that rules for analyzing spontaneously
spoken sentences can hardly be foreseen by a grammarian.
We believe that a system that learns from examples will ease the
work of a designer of text or speech understanding system giving
the possibility of analyzing big corpora of sentences.
The question remains on how to collect those corpora,
which kind of annotation is needed, and what is the amount of
manual work that has to be carried on.\\

The basis of the work exposed in this paper is a semantic translator,
called 
{\it CHRONUS.}~\footnote{{\it CHRONUS} stands for Conceptual Hidden
Representation of Natural Unconstrained Speech.}
CHRONUS is based on a stochastic
representation of conceptual entities resulting from the formalization
of the speech/text understanding problem as a communication problem.
The paper is structured as follows. In section~\ref{sect:model} we formalize
the language understanding problem and we propose an algorithm based on
the maximum a posteriori decoding. In section~\ref{sect:implement} we
explain how the described algorithm for conceptual decoding can be part
of a complete understanding system and we give a short description of all the
modules that were implemented for an information retrieval application.
In section~\ref{sect:practice} we discuss experimental performance of
the system as well as issues related to the training of the conceptual
decoder. Finally in section~\ref{sect:conclusion} we conclude the
paper with a discussion on the open problems and the future developments
of the proposed learning paradigm.

\section{Formalization of the Language Understanding Problem
\label{sect:model}}
In this section we propose a formalization of the language
understanding problem in terms of the noisy channel 
paradigm. This paradigm has been introduced for formalizing the
general speech recognition problem~\cite{jelinek-76} and constitutes
a basis for most of the current working speech recognizers. Recently,
a version of the paradigm was introduced for formalizing the problem
of automatic translation between two languages~\cite{brown-90}.
The problem of translating between two languages has the same flavor of
the problem of understanding a language~\cite{prieto-91}. In the
former, both the input and the output are natural languages, while in the
latter the output language is a formal semantic language apt to
represent meaning. 

The first assumption we make is that the meaning of a sentence can be expressed by
a sequence of basic units $ {\bf {\cal M}} ~=~  \mu_{1}, 
\mu_{2} , \ldots , \mu_{N_{M}} $ and
that there is a {\em sequential correspondence} between each $ \mu_{j}$ and
a subsequence of the acoustic observation $ {\bf A}  ~=~ a_{1}, a_{2}
\ldots a_{N_{A}} $, so that we could actually segment the acoustic signal
into consecutive portions, each one of them corresponding to a phrase that
express a particular $\mu_{i}$.
The second assumption consists in thinking of the acoustic representation of
an utterance as a version of the original sequence of meaning units corrupted
by a noisy channel whose characteristics are generally unknown.
Thus, the problem of understanding a sentence can be expressed in this
terms: given that we observed a sequence of acoustic measurements $ {\bf A} $
we want to find which semantic message $ {\bf {\cal M}} $
most likely produced it, namely the one for which
the a posteriori probability
$P( {\bf {\cal M}} \mid \bf A )$ is maximum. Hence the problem of understanding 
a sentence is 
reduced to that of maximum a posteriori probability decoding (MAP).

For the actual implementation of this idea we need 
to represent the meaning of a sentence as a sequence
of basic units. A simple choice consists in defining a
unit of meaning as a {\it keyword/value} pair
$  m_{j} = ( k_{j}, v_{j})$, where
$k_{j}$, is a conceptual category (i.e. a
{\em concept} like for instance {\em origin of a
flight, destination, meal})  and $ v_{j} $ is the
{\em value} with which $ k_{j}$ is instantiated in the actual sentence
(e.g. {\em Boston, San Francisco, breakfast}). Given a certain
application domain we can define a concept dictionary $\Gamma $
and for each concept $ \gamma_{j} \in \Gamma $ we can define a set
of values  $ \Upsilon^{j} = \{ \varphi{1}^{j}
, \varphi{2}^{j} , \ldots , \varphi{N_{v}^{j}} \} $.
Examples of meaning representation for
phrases in the airline information domain are given in Table~\ref{tab:mean-examples}.

\begin{table}
\centering
\begin{tabular}{|p{2.5in}|p{3.5in}|}
\hline
{\it SHOW ME THE FLIGHTS TO BOSTON} & ({\sf question},{\sf display}) ({\sf subject},{\sf flight}) ({\sf destin},{\sf BBOS})
\\ \hline
{\it HOW MUCH IS THE PRICE OF THE FLIGHT FROM ATLANTA} & 
({\sf question},{\sf display}) ({\sf subject},{\sf fare}) ({\sf destin},{\sf MATL}) \\ \hline
{\it IS BREAKFAST SERVED ON THE FLIGHT?} & ({\sf question},{\sf yes-no}) ({\sf subject},{\sf breakfast}) \\ \hline
\end{tabular}
\caption{Example of keyword/pair representations of simple phrases within
the ATIS domain.\label{tab:mean-examples}}
\end{table}

For information retrieval applications, the number of concepts
$\Gamma$ is relatively small (about 50, in the ATIS application),
while the dictionary of
concept values $ \Upsilon $ can be relatively large (consider for
instance all the possible flight numbers in the airline reservation domain). Moreover,
the limited amount of training data available
at the time we developed the system (a few thousand sentences) suggested to keep 
the number of model parameters relatively small.

Therefore, in the decoding process,  we considered only concepts $k_{j} \in \Gamma $;
the concept values $v_{j} \in \Upsilon$ are derived
through pattern matching functions at a later stage of processing.

In the remaining of the section we will use the following notations:
$ {\bf W} = w_{1} , \ldots , w_{N_{W}} $ is the sequence of words
actually uttered in the sentence represented by the acoustic observation
${\bf A}$ and
$ {\bf  C } =  c_{1}  , \ldots ,  c_{N_{W}} , ~ ~ c_{i} \in \Gamma $ 
is the corresponding sequence of concepts labels. A consecutive set of
words $ w_{i}, \ldots , w_{i+k} $ labeled by the same concept label, constitutes
a phrase expressing a concept, thus
defining a segmentation of the
sentence into concepts, referred to, in the following, as {\em
conceptual segmentation}.

Hence, according to the maximum a posteriori decoding criterion, given
the sequence of acoustic observations ${\bf A}$, we want to find
the sequence of conceptual labels ${\bf \tilde{C}}$ and the sequence
of words $ {\bf \tilde{W}} $ that maximize
the a posteriori probability $ P( \bf W , \bf C \mid \bf A ) $, namely
\begin{equation}
P( {\bf \tilde{W}} , {\bf \tilde{C}} \mid {\bf A} )~=~
\max_{ {\bf W } , {\bf C }}
~ P( {\bf W }, {\bf C} \mid {\bf A} ).
\label{eq:1}
\end{equation}
Using the Bayes inversion formula, the conditional probability can 
be rewritten as:
\begin{equation}
P( {\bf W} , {\bf C}  \mid {\bf A} )~=~\frac{P( {\bf A} \mid {\bf W} , {\bf C} )
P(  {\bf W} \mid  {\bf C} )P( {\bf C} )}{P( {\bf A} )},
\label{eq:2}
\end{equation}
and since $P( \bf A )$ is constant:
\begin{equation}
\arg \max_{ {\bf W } , {\bf C }}
~ P( {\bf W} , {\bf C} \mid {\bf A} )~=~
\arg \max_{ {\bf W } , {\bf C }} ~ P( {\bf A} \mid {\bf W} , {\bf C} ) 
P( {\bf W} \mid {\bf C} ) P( {\bf C} )
\label{eq:3}
\end{equation}
In this formula $ P (\bf C ) $ represents the a-priori probability of the
sequence of concepts, $ P( \bf W \mid \bf C ) $ is the probability of the
sentence (intended as a sequence of words), given that the sentence
conveys the sequence of concepts $\bf C$, and
$ P( \bf A \mid \bf W ,\bf C ) $ is the acoustic model. The three
components of the conditional probability in~\ref{eq:2} can be
thought of also as a semantic (probability of meanings), 
syntactic (probability of words given a meaning), and acoustic component respectively.
We can then reasonably assume that the acoustic representation of a word is
independent from the conceptual relation it belongs to, hence:
\begin{equation}
P( {\bf A} \mid {\bf W} , {\bf C} ) ~ = ~ P( {\bf A} \mid {\bf W} ),
\end{equation}
and this is the criterion that is usually maximized in stochastic based
speech recognizers, for instance those using hidden Markov modeling
~\cite{jelinek-76}~\cite{rabiner-93} for the acoustic/phonetic decoding.
Both the syntactic and the semantic probabilities can be
rewritten as:
\begin{equation}
P ( {\bf W} \mid {\bf C} ) ~ = ~ P ( w_{1} ) \prod_{i=2}^{N_{W} }
P ( w_{i} \mid w_{i-1} , \ldots , w_{1} , \bf C )
\label{eq:synt}
\end{equation}
\begin{equation}
P ( {\bf C} ) ~ = ~ P ( c_{1} ) \prod_{i=2}^{N_{W}}
P ( c_{i} \mid c_{i-1} , \ldots , c_{1} ).
\label{eq:sem}
\end{equation}
We assume that
the probability of a word $w_{i}$ given all
the previous words in the sentence and the sequence of concepts $ \bf C $,
depends only upon the $ n $ most recent
words and the concept $ c_{i} $ that the word $w_{i}$
contributes to express. Under this assumption, equation~\ref{eq:synt}
can be rewritten in terms of the {\em n-gram concept conditional}
word probabilities:
\begin{equation}
P ( w_{i} \mid w_{i-1} , \ldots , w_{1} , {\bf C} ) =
P ( w_{i} \mid w_{i-1} , \ldots , w_{i-n+1} ,  c_{i} )
\label{eq:synt.mm}
\end{equation}
Analogously, the sequence of concept labels $ \bf C $ can be regarded as an
$m-th$ order Markov process under the assumption:
\begin{equation}
P ( c_{i} \mid c_{i-1} , \ldots , c_{1} ) ~=~
P ( c_{i} \mid c_{i-1} , \ldots , c_{i-m} )
\label{eq:sem.mm}
\end{equation}
For $n = m = 1$ we can represent the probabilities in
equations~\ref{eq:synt.mm} and~\ref{eq:sem.mm} as a first order
hidden Markov model, whose states represent the conceptual
labels and whose observations are the words. 

Given the representation of the conceptual structure as a traditional
HMM, the decoding of the conceptual content of a sentence
can be carried out with the Viterbi algorithm.
If the input is a text sentence,  Viterbi decoding is used for
finding the sequence of states $\bf \tilde{C}$ such that:
\begin{equation}
P ( \bf W , \bf \tilde{C} ) ~=~ \max_{\bf C} P ( \bf W , \bf C ).
\end{equation}
If the input is speech, we want to find $\bf \tilde{W}$ and
$\bf \tilde{C}$ that maximize equation~\ref{eq:3}. In principle, the
solution to this problem can be found by substituting 
each state of the conceptual HMM with a finite state network
representing the corresponding bigram structure, and by substituting
each word with its corresponding acoustic HMM. The integrated network,
obtained in this way, can be used for decoding the input
speech with Viterbi algorithm. 

\section{Implementation of a Speech  Understanding System \label{sect:implement}}
For putting into practice the idea of MAP conceptual decoding,
or at least to show its effectiveness, one needs a corpus of
training examples (sentences and their correspondent meaning) and
a test set with a criterion for validating the results. A
relatively large corpus expressively designed for speech
understanding (but not for learning) is being developed within the
DARPA ATIS project~\cite{price-90}.
ATIS stands for Air Travel Information System and the 
task is
built around a subset of the OAG (Official Airline Guide)
database, including 10 American cities. A corpus of spontaneous
sentences is being collected and annotated
by different sites~\cite{madcow-92}. The corpus is collected through
a Wizard of Oz paradigm.
Each subject is given a scenario
and a travel planning problem to solve. The subjects are requested
to solve the problem by interacting 
with a machine (that is actually a human {\em wizard}).
The partial and the final responses of the machine are
presented to the subjects via a display or a speech
synthesizer.
The sentences uttered by the subjects are recorded,
transcribed and annotated carefully. 

Although the ATIS corpus may not be the best corpus for testing
a semantic learning paradigm, it is readily available and
it includes some kind of meaning annotation that can be
indirectly used for our purpose.
In this section we will give details 
about the design and test of a complete speech understanding
system for the ATIS task.

\begin{figure}[ht]
\centerline{\psfig{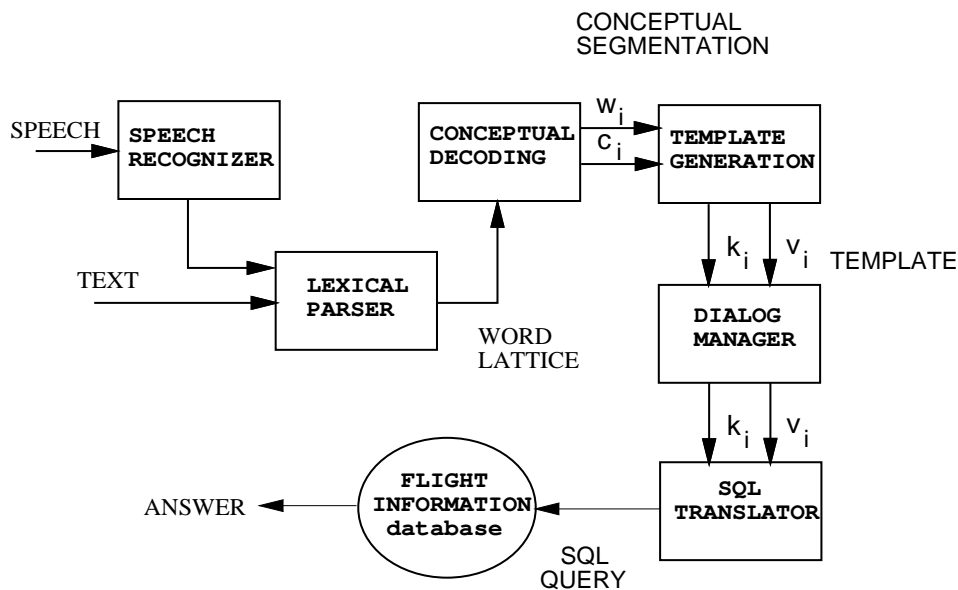}}
\caption{Block diagram of the understanding system \label{fig:block-d}}
\end{figure}

A block diagram of the speech understanding system is 
shown in Fig.~\ref{fig:block-d}. 

The system can work both
from speech and text input. The {\em lexical parser} 
preprocesses sentence transcription in order
to assign words to word categories (like numbers and
acronyms). The {\em conceptual decoding} provides a segmentation
of the sentence into phrases associated with conceptual units according to the theory
explained in chapter~\ref{sect:model}. The {\em template
generator} assigns to each concept the proper value,
while the {\em dialog manager} takes care of keeping the history
of the dialog and including in the current template the missing
information.
Finally the {\em SQL~\footnote{SQL, Structured Query Language, is a
high level language for accessing data in a relational database.}
translator} generates the query. The appropriate information, stored
in a relational database, is then retrieved and displayed.

\subsection{The Conceptual Decoding}
\begin{figure}[ht]
\centerline{\psfig{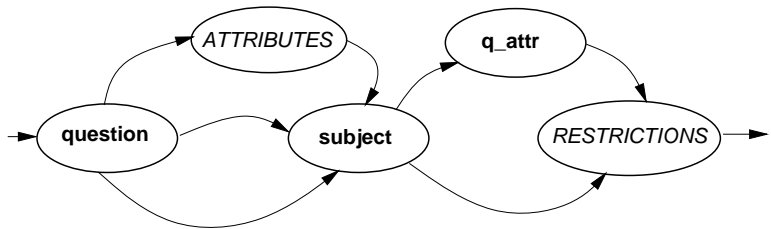}}
\caption{Typical structure of a query \label{fig:sentstr}}
\end{figure}

Although MAP decoding of concepts is purely based on semantics,
and the final structure of the conceptual model is in principle
completely data driven,
a certain amount of structure can be imposed during the design
of the concept dictionary $ \Gamma $. Imposing a certain structure
to the model certainly alleviates the problems of the locality
of the model and that of concept embedding, as explained
later.
For designing the concept dictionary,
we had in mind the
typical structure of a query that is
represented as in Fig.~\ref{fig:sentstr}.

In a typical sentence there is a phrase that generally represent
the {\it question} , then a {\it subject} and finally a {\it restriction} on
the query. For example, in the sentence:
\begin{quote}
{\it SHOW ME THE FLIGHTS TO SAN FRANCISCO IN THE MORNING}
\end{quote}
the question phrase is {\it SHOW ME}, the subject is {\it THE
FLIGHTS} and the restriction is {\it TO SAN FRANCISCO IN THE
MORNING}. The same sentence could be rephrased as:
\begin{quote}
{\it SHOW ME THE MORNING FLIGHTS TO SAN FRANCISCO}
\end{quote}
{\it THE MORNING} in this sentence plays exactly the
same semantic role as the phrase {\it IN THE MORNING} of the
previous sentence, but it has a different syntactic connotation.
Therefore when a restriction is placed in front of the subject of the
question we call it an {\em attribute}. Many different concepts
can play both the role of restriction and attribute, hence we
represent them by separate entities. For instance there is a
{\bf fare} concept and an {\bf a\_fare} concept that are
semantically indistinguishable but with the syntactic role
of restriction and attribute respectively. Giving some syntactic
connotations to the concepts helps to have better and sharper
stochastic models. Without distinguishing between attribute and
restriction concepts, the transition probabilities tend to
be more uniform and the concept dependent bigram language models
tend to be larger since the expressions used for an attribute concept
are often different than those used for a restriction concept ({\it
IN THE MORNING} vs. {\it THE MORNING}).

Another problem is that of concept embedding. The basic assumption
we made in section~\ref{sect:model}, namely that the meaning can
be expressed as a sequence of {\em meaning units} is not generally
true~\footnote{Semantics can be represented, in general, by a tree
~\cite{minski-75} or by a network~\cite{sowa-84}. The assumption made
here is that the semantics represented by the sentences in the
application we are considering is simple enough and can be represented
with a flat structure like a sequence of symbols}
Take for instance the following sentence:
\begin{quote}
WHAT TYPE OF ECONOMY FARE COULD I GET FROM SAN FRANCISCO
TO DALLAS 
\end{quote}
The concept of {\bf question} is represented by two different
phrases, namely {\it WHAT TYPE OF} and {\it COULD I GET}. Other
concepts are embedded in between. Since we are using a flat
representation for the meaning (a sequence rather than a tree)
one of the possible solution for coping with this problem is
to define additional symbols to account for separate
phrases in which a single concept can be broken into. For example,
in the previous sentence, we introduce a {\bf q\_attr} (question
attribute) concept, and the corresponding conceptual segmentation
will be as follows:

\begin{center}
\begin{tabular}{ll}
 {\sf question:} & {\it WHAT TYPE OF} \\
 {\sf a\_fare:} & {\it ECONOMY} \\
 {\sf subject:} & {\it FARE} \\
 {\sf q\_attr:} & {\it COULD I GET} \\
 {\sf origin:} & {\it FROM SAN FRANCISCO} \\
 {\sf destin:} & {\it TO DALLAS} \\
\end{tabular}
\end{center}

There are other special concepts like {\sf dummy}, that accounts
for phrases that do not carry information that is relevant for the
application (e.g. courtesy forms, etc.), and {\sf and} that
represent conjunctives (e.g. and, or, also, etc.). The 
concept dictionary used in this implementation counts a total of
48 concepts.

\subsection{The Lexical Parser}

When dealing with a {\em speech} understanding system 
there are different issues related to lexicon that may
not be encountered in {\em text} understanding.
\begin{itemize}
\item Although in both systems the vocabulary is 
generally limited
in size, the limitations in a speech system are more severe than
in a text system. 
\item A text system can generally deal with unknown (out
of vocabulary) words. Once
an unknown word is spotted, the system can take some action, like
for instance asking the user to
rephrase the sentence using a different word. In a speech 
recognition system an unknown word 
cannot generally be detected reliably by the system; 
it is generally confused and substituted for a known word.
\item Numbers and acronyms are generally written in an unambiguous
form but they can be uttered in different ways and allow to ambiguous 
interpretations.
For instance, the acronym DC10 can be uttered as {\it D C TEN}
or {\it D C ONE ZERO}, or {\it D C ONE OW} and can be interpreted
as DC10, or D C10, or DC 10, etc.
\end{itemize}

The lexical parser analyzes the input transcription and generates
a lattice of word categories called {\em superwords}.
A superword can be one of the following:
\begin{itemize}
\item[a.]
a word, like ABOUT, MONTH, RETURN, etc.
\item[b.]
a word with optional morphological inflections, like AIRFARE(S), DAY(S),
ADVANCE(D), etc. (e.g. there is no
distinction between AIRFARE and AIRFARES; both words are represented
by the super-word AIRFARE(S)).
\item[c.]
a grammar, represented by a finite state automaton (FSA); for instance, the
grammar for natural numbers (e.g. THIRTY SEVEN), the grammar for 
airport acronyms (e.g. D F W ), the grammar for airport names
(e.g. SAN FRANCISCO INTERNATIONAL AIRPORT), etc.
Each sequence recognized by an FSA is characterized by
the corresponding FSA identifier and a normalized form of
the compound word (e.g. THIRTY SEVEN is represented
as ((number)37), D F W as ((airport)DFW)).  As far as the stochastic
model is concerned, two words with the same FSA identifier are represented
by the same super-word.
\end{itemize}

When there is ambiguity in deciding which superword to assign to 
a sequence of words, the lexical parser generates all the possible
interpretations and arranges them in a lattice~\footnote{
It is quite straightforward to modify the Viterbi algorithm for
performing the decoding from a lattice rather than from a sequence
of observations. A discussions on this subject can be found in
~\cite{fissore-89}}.

Besides, articles (i.e. {\it THE} and {\it A} ) are deleted by the 
lexical parser, hence they play no role in the conceptual decoding process.
Concept conditional language models are thus estimated among
content words only.
The reason for doing this is that 
in this particular application there is no
relevant information carried by an article in front of a
noun or an adjective.
Besides, articles (like other short function
words) can be easily misrecognized or deleted by the speech
recognizer. Hence, due to the locality of the language model
(i.e. bigrams of words), many bigrams will carry the
probability of a noun or an adjective preceded by an article,
missing the more important correlation to the preceding content word.\\

Using super-words reduces the number of parameters to be estimated and increases
the robustness of the system. The probabilities associated to a given
super-word are shared between all the words that are represented
by the same super-word. This has the effect of allowing the system
to generalize the statistics gathered in the training phase to all
the words belonging to the same super-word. This is shown in an
experiment where we implemented the conceptual decoding in two
different situations. 
In the first implementation there was no super lexicon, 
the vocabulary consisted of 501 words, including three word classes
(i.e. numbers, city names, and acronyms), and the articles were
accounted for in the bigrams. In this case, the input to the system
was a textual transcription of each utterance, where the numbers
and the acronyms were unambiguously represented (e.g. {\it What's
the fare for US-AIR 4393}). In the second experiment a
super-lexicon including 753 words and 18 grammars was used. In 
this case the input to the system was a speech transcription
of each utterance.
It has to be noticed that although the lexicon in the second experiment 
was larger than the lexicon in the first experiment,
the word coverage over the test set was the same in both cases.
The test set consisted of 148 context independent sentences (included in the
official February 1991 test set~\cite{pallett-91}). The 148 sentences
were hand segmented into concepts.
The set of sentences corresponded
to an overall number of 713 concepts. 
The performance were assessed based on the hand segmentation. The
percentage of correct concepts (when the conceptual segmentation agreed
with the one provided by hand) and of correct sentences
(sentence for which all the included concepts are correctly decoded
and segmented) is reported in Table~\ref{tab:results1}.

\begin{table}
\centering
\begin{tabular}{|c|c|c|} \hline
{\bf Experiment} & {\bf \% of correct } & { \bf \% of correct } \\
{\bf Description} & {\bf concepts} & {\bf sentences} \\ \hline
no super-lexicon & 94.1 & 80.4 \\ \hline
super-lexicon & 96.8 & 88.5 \\ \hline
\end{tabular}
\caption{Effect of the super-lexicon in conceptual 
decoding \label{tab:results1}}
\end{table}

\subsection{Interfacing with a Speech Recognizer}
The most natural way of interfacing the conceptual decoder with a
speech recognizer is by implementing the maximization of equation~\ref{eq:3}.
This requires to implement a decoder that explores a network obtained
by explicitly instantiating acoustic HMMs~\cite{lee-90} 
representing  words
(or superwords) of the vocabulary for any concept. For a
task like ATIS the dimension of the resulting network can be rather large. In
theory, if there are 50 conceptual states and about 1,000 words, each
one of them represented by a HMM with an average number of 15 states,
the overall network is bound by a total number of 750,000 acoustic HMM
states, with a number of connections of the order of 50,000,000 (each
conceptual conditional bigram model is represented by $1000 \times 1000$
connections). Of course not all the bigrams are observed or even possible
in each state. If only those words and bigrams that were observed during
the training are represented in a conceptual state, a more reasonable
model can be obtained. In an experimental version of CHRONUS we estimated
an integrated model with a total of 2400 HMM word models (corresponding
to about 36,000 HMM acoustic states) and nearly 46,000 connections. This
size of the model can be easily managed by a beam search 
recognizer~\cite{pieraccini-90}. The problem in using such a model is
that while it constitute a reasonably good model for decoding the semantic
message of a sentence, the limited amount of training data used
for its estimation makes it a quite coarse model for constraining
the speech recognition process. When bigrams of words that were not
observed in the training data are actually uttered, the recognizer
is forced to substitute them for known bigrams. Hence the recognition errors are
propagated along the sentence, resulting in relatively poor recognition performance.

Smoothing techniques can be applied for estimating the
probability of unobserved bigrams, like for instance methods relying
on the Good-Turing estimation of probabilities~\cite{katz-87}. This will
increase the complexity of the model by allowing all the possible bigrams
in each state. However a factorization of the
maximization of equation~\ref{eq:3}~\cite{pieraccini-91-2} can still
lead to reasonably good results at an acceptable complexity. Hence
several solutions could be implemented, like best first coupling (the
best first recognized sentence is given to the conceptual decoding),
N-best coupling~\cite{ostendorf-91} and word lattice coupling~\cite{fissore-90}.

\subsection{The Template Generator}
The goal of the template generator module is that of analyzing
the conceptual segmentation and generating the final representation
of the meaning (i.e. the {\em template}) by supplying the correct
value to each concepts detected during the decoding stage.
For every relevant concept a look-up table was built for performing the
mapping between phrase templates and conceptual values. The look-up table
associates keywords or short phrases to concept values. Values are
then assigned to the decoded concepts according to the result of a
pattern matching procedure with the keyword stored in the appropriate
tables.
The values that are eventually associated to the decoded concepts
belong to different categories. They can be actual
database items (e.g. {\bf Boston, American airlines, breakfast}), database
attributes, (e.g. {\bf flight, stop, meal}), logic values 
(e.g. {\bf null, not null}) or
operators, (e.g. {\bf minimum, maximum}). The design of pattern
matching tables is
still manual. 
More details on the template generator can be found
in~\cite{tzoukermann-92}.

\subsection{The Dialog Manager}
The {\em dialog manager} implements the function of keeping the dialog
history and allowing the resolution of anaphoric and elliptical
sentences. The simple strategy implemented
in our system consists in keeping a current {\em context template} with
all the information that has been used for specifying the actual query.
When a new sentence is presented to the system, the dialog manager tries
to merge the current template with the current context template, in order
to get the missing information. The merging of the two templates 
follows
{\em application specific} rules. For instance,
when a concept is mentioned in a new
sentence, with a different value than the corresponding concept in 
the context template, all concepts in the context template
at a lower hierarchical level are deleted. This assumes a predefined
hierarchy of the concepts.
The origin and the destination of a
flight have the highest level in the hierarchy. Then, when either
the origin or the destination are different from those specified in
the context template, the context template is deleted and a new
context is started. 
This strategy, although very simple, has proven to be effective in the
majority of sentences of this task. 

\subsection{The SQL translator}

The last part of the interpretation process, namely the access to the
required information, is implemented through a
translator that dynamically generates the SQL query in order to 
retrieve the data.
The template produced by the template generator
is processed according to the value of its {\sf subject} concept. If a 
{\sf subject} concept
is not found in the template a default
subject is used.
The subject of the query is used for selecting the
right table of the database. If there is more than one subject or
the subject is not directly related to a particular table, a link
function is invoked in order to perform the correct joins. Once
the table (or a joint table) is selected, the rest of the template tokens
are interpreted accordingly.\\

\section{Putting it into Practice \label{sect:practice} }
Assessing the performance of a language understanding
system is still an open problem mainly because the concept of
{\em correct answer} is generally ambiguous and must be based on
defined conventions that are not task independent.
The DARPA community agreed
upon scoring answers by comparison with given reference answers
that are produced for each valid sentence of the corpus. The answers
can be made up of data extracted from the flight information database,
numbers, or logical values (yes/no).
In case of ambiguity of the question, multiple reference answers
are given. 
Of course the problem of the definition of a {\em correct} answer
still remains.
For instance, for a question like
\begin{quote}
{\it SHOW THE LATE EVENING FLIGHTS BETWEEN BOSTON AND DALLAS}
\end{quote}
the correctness of the answer depends upon the conventional definition
of {\em late evening}. Then, once a time interval has been defined
for late evening,
it is still not clear what is the information to be listed. It could
be the airline and flight number of each flight, but it
could also include the departure time, the arrival time, 
the fare, and so on. A special committee within the DARPA
community agreed upon a certain number of rules, called
{\em principles of interpretation}~\cite{madcow-92}, that should rule
the majority of cases. Besides, it was also agreed on using
two reference answers, namely a {\em minimal} and a {\em maximal}
reference answer. 
An answer is thus considered correct
if it contains all the information included in
the minimal reference answer and no more than the information included in the
maximal reference answer.
\subsection{Experimental Performance}
The described system was tested on a set 
of 687 sentences called February 92 test set~\cite{pallett-92}.
For increasing the robustness we provided the system
with a simple rejection criterion.
The rejection heuristic is based on the measure of success in 
the operation of template generation (how many decoded concepts
are successfully matched to a value), although more sophisticated
heuristics can be developed. The results~\cite{pieraccini-92-2} 
on the complete test set,
from text input, account for 68\% of correct answers, 18\% wrong
answers and 14\% rejects.  When the system was coupled with a speech
recognizer through the best first hypothesis, the performance 
dropped to 52\% of correct answers, 26\% wrong answers and 22\%
rejects.
However, the contribution of the
different modules to the overall error rate is far more interesting
than it absolute value.
All the 121 sentences that produced a wrong answer (in the
text understanding experiment)  were carefully analyzed
and the errors were classified according to Table~\ref{tab:nl.analysis}.
\begin{table}
\centering
\begin{tabular}{|c|c|}
\hline
{\bf Error type} & {\bf Number of} \\
& {\bf Sentences} \\ \hline
Conceptual decoding & 30 \\ \hline
Template generation & 19 \\ \hline
Dialog manager & 20 \\ \hline
SQL translator & 52 \\ \hline
\end{tabular}
\caption{Analysis of the errors for the NL ATIS February 92 
test \label{tab:nl.analysis}}
\end{table}
From this analysis it results that most of the errors are due to
the parts of the system that are not trained. The template generator
errors reflect a lack of entries in the look-up tables. The dialog manager
errors are due to the fact the the simple strategy for merging the
context and the current template should be refined with more sophisticated
rules. The SQL translator should be an error-free module. Its only
function is that of translating between two different representation
in a deterministic fashion. However, in this test, the SQL module
faced two kinds of problems.
The first is that the interpretation rules used
for generating the answers were not exactly the same ones used 
for the official test, and this accounts for roughly half of the
errors. The other half of the error is due to the limited
power of the template representation, and this will be discussed
in section~\ref{sect:conclusion}.
\subsection{Training the Conceptual Model}
\subsubsection{Smoothing of Bigram Models}
The conceptual model, as explained in section~\ref{sect:model}, is defined by
two sets of probabilities, namely the {\em concept conditional
bigrams} $ P ( w_{i} \mid w_{i-1} , c_{i} ) $ and the {\em concept
transition probabilities} $ P ( c_{i} \mid c_{i-1} ) $. In the first
experiments these probabilities were estimated using a set of
532 sentences whose conceptual segmentation was provided by hand.
The accuracy of the system in the experiments carried out using 
the model estimated with such a
small training set, although surprisingly high~\cite{pieraccini-91}, shows a definite
lack of training data. Smoothing the estimated model probability
provides  an increase of the performance. 
The knowledge of the task can be introduced through a
{\em supervised smoothing} of the concept conditional bigrams.
The supervised smoothing is based on the observation that,
given a concept,  there are several words that carry
the same meaning. For instance, for the concept {\bf origin},
the words
\begin{center}
DEPART(S) LEAVE(S) ARRIVE(S)
\end{center}
can be considered as synonyms, and can be interchanged in sentences
such as:
\begin{center}
{\it THE FLIGHT THAT DEPART(S) FROM DALLAS} \\
{\it THE FLIGHT THAT LEAVE(S) FROM DALLAS} \\
{\it THE FLIGHT THAT ARRIVE(S) FROM DALLAS.}
\end{center}
A number of groups of synonyms were manually compiled for each concept. The
occurrence frequencies inside a group were equally shared among the constituting
words, giving the same bigram probability for synonymous words.\\

\subsubsection{Using a Larger Training Corpus}
If one wants to use a larger corpus than the initial handlabeled
few hundred sentences and wants to avoid an intensive hand segmentation
labor, one has to capitalize on all the possible information associated
to the sentences in the corpus. Unfortunately, when the corpus is not
expressively designed for learning, like the ATIS corpus, the information
needed may not be readily available. In the remaining of this section we
analyze solutions that, although particularly devised for
ATIS, could be generalized to other corpora and constitute a guideline
for the design of new corpora.

A training token consists of a sentence and its associated meaning.
The meaning of sentences in the ATIS corpus is not available
in a declarative form. Instead, each sentence is associated with 
the {\em action} resulting from the {\em interpretation} of the 
meaning, namely the correct answer. One way of using this information
for avoiding the handlabeling and segmentation of all the sentences in the
corpus consists in creating a training loop 
in which the provided correct answer serves the purpose of a feedback signal.
In the training loop all the
available sentences are analyzed by the understanding system obtained
with an initial estimate of the conceptual model parameters. The
answers are then compared to the reference answers and the sentences
are divided into two classes. The {\em correct} sentences, for
which we assume that the conceptual segmentation obtained with 
the current model is correct, and the {\em problem sentences}.
Then the segmentation of the correct sentences is used for
reestimating the model parameters, and the procedure is
repeated again. The procedure can be repeated until it converges
to a stable number of correct answers. Eventually, the remaining {\em problem
sentences} are corrected by hand and included in the set of correct
sentences for a final iteration of the training algorithm. This
procedure proved effective for reducing the amount of handlabeling.
In the experiment described in~\cite{pieraccini-92} we showed that
the performance increase obtained with the described training loop,
without any kind of supervision (the remaining {\em problem}
sentences were excluded from the training corpus)
is equivalent to that obtained with the supervised smoothing. This
means that the training loop, although is not able to learn radically
new expressions or new concepts, is able to reinforce the acquired
knowledge and to {\em infer} the meaning of semantically equivalent 
words.
In a set of 4500 sentences the training loop automatically classified almost 80\% of the
sentences, leaving the remaining 20\% to the manual segmentation.\\

\subsubsection{The Sequential Correspondence Assumption \label{sect:pseudoE}}
In section~\ref{sect:model} we based our formalization of the
speech understanding problem on the assumption that there is
a sequential correspondence between the representation of a sentence 
(words or acoustic measurements)
and the corresponding representation of meaning.
This assumption is not generally true for any translation (semantic
or not) task. An interesting example (reported in~\cite{oncina-93})
of a task where there is
no sequential correspondence between a message and its
semantic representation, is that
of roman numbers (e.g.  I, XXIV, XCIX) and their correspondent
decimal representation (e.g. 1, 24, 99). Fortunately, in a natural language
understanding task, we may have the freedom of choosing the
semantic representation, like we did in the implementation of
CHRONUS explained above. But in general, if we are dealing with
a large corpus of sentences that have not been expressively designed
for the purpose of learning a semantic translator, and we
would like to take advantage of some kind of semantic annotation
already available, we may have to face
the problem of the not sequentiality of the representation. For instance, in the
ATIS corpus, each sentences is associated
with the intermediate representations used by the annotators for obtaining
the reference correct answers. In fact the annotators rephrase each
valid sentence in an artificial language that is a very restricted form
of English. This {\em pseudo-English} rephrasing 
(called {\em win} or {\em wizard input}) constitute the input of
a parser, called NLparse~\cite{hemphill-90}, that unambiguously
generates the SQL query.
For instance, for a sentence like:
\begin{quote}
{\em I'D LIKE TO FIND THE CHEAPEST FLIGHT FROM WASHINGTON D C TO ATLANTA}
\end{quote}
The {\em win} rephrasing is:
\begin{quote}
{\em List cheapest one direction flights from Washington and to Atlanta}
\end{quote}
and the corresponding associated SQL statement is:
\begin{quote}
\tiny
(SELECT DISTINCT flight.flight\_id FROM flight WHERE (flight.flight\_id IN 
(SELECT flight\_fare.flight\_id FROM flight\_fare WHERE flight\_fare.fare\_id IN
(SELECT fare.fare\_id FROM fare WHERE fare.one\_direction\_cost =
(SELECT MIN ( fare . one\_direction\_cost ) FROM fare WHERE fare.fare\_id IN
(SELECT flight\_fare.fare\_id FROM flight\_fare WHERE flight\_fare.flight\_id IN
(SELECT flight.flight\_id FROM flight WHERE (flight.from\_airport IN
(SELECT airport\_service.airport\_code FROM airport\_service WHERE airport\_service.city\_code IN
(SELECT city.city\_code FROM city WHERE city.city\_name = 'WASHINGTON' ))
AND flight.to\_airport IN
(SELECT airport\_service.airport\_code FROM airport\_service WHERE airport\_service.city\_code IN 
(SELECT city.city\_code FROM city WHERE city.city\_name = 'ATLANTA' ))))))))
AND (flight.from\_airport IN
(SELECT airport\_service.airport\_code FROM airport\_service WHERE airport\_service.city\_code IN
(SELECT city.city\_code FROM city WHERE city.city\_name = 'WASHINGTON' ))
AND flight.to\_airport IN
(SELECT airport\_service.airport\_code FROM airport\_service WHERE airport\_service.city\_code IN
(SELECT city.city\_code FROM city WHERE city.city\_name = 'ATLANTA' )))));.
\end{quote}
Both the SQL query and the {\em win} sentence can be considered
semantic representations of the original sentence. In fact the SQL
query is the final target of the understanding system and can
be unequivocally obtained from the {\em win} sentence through
an existing parser. Obviously the sequential correspondence assumption
is strongly violated for the SQL representation. However a sequential
correspondence can be easily found between the pseudo-English {\em win}
sentence and the original message, at least for the shown examples.
Since all the valid sentences in the ATIS corpus have a {\em win}
annotation, the pseudo-English language can be thought of as an
alternate candidate for the meaning representation in our learning
framework. Using {\em win}
for representing the meaning may lead to two different solutions.
In the first we can think of developing a system that learns how to
translate natural language sentences into pseudo-English sentences and
then use the existing parser for generating the SQL query. In the
second solution each {\em win} sentence in the corpus can be
translated in the corresponding conceptual representation used
for CHRONUS. This translation is unambiguous ({\em win} is an 
unambiguous artificial language by definition). A parser can be
easily designed for performing the translation or, simply use CHRONUS
itself for performing the translation\footnote{{\em win} is a subset of 
natural English. Only a little adaptation was needed for developing
a {\em win} translator based on the existing CHRONUS}.
Unfortunately also for the {\em win} representation, the sequential
correspondence assumption is violated for a good percentage of the sentences in
the corpus. A typical example is constituted by the following sentence:
\begin{quote}
{\em COULD YOU PLEASE GIVE ME INFORMATION CONCERNING AMERICAN AIRLINES 
A FLIGHT FROM WASHINGTON D C TO PHILADELPHIA THE EARLIEST 
ONE IN THE MORNING AS POSSIBLE}
\end{quote}
whose corresponding {\em win} annotation is:
\begin{quote}
{\em List earliest morning flights from Washington and to Philadelphia and American}.
\end{quote}
The problem of reordering the words of {\em win} representation for
aligning it with the original sentence is a complex problem that
cannot be solved optimally. Suboptimal solutions with satisfactory
perfomance can be developed based on effective
heuristics.  We will not discuss the details of how the reordering
can be put into practice. Rather we want to 
emphasize the fact that an iterative algorithm based on a model similar to
that explained in section~\ref{sect:model}
led to almost 91\% correct alignments between English sentences and 
corresponding {\em win}
representations on a corpus of 2863 sentences. With additional
refinements this technique can be used, integrated in the
training loop, for automatically processing the training corpus of 
the conceptual model.

\section{Discussion and Conclusions \label{sect:conclusion}}

In this paper we propose a new paradigm for language understanding
based on a stochastic representation of semantic entities called
concepts.
An interesting way of looking at the language understanding paradigm
is in term of a language translation system.
The first block in  Fig.~\ref{fig:transl} translates a sentence
in natural language ({\em N-L}) into a sentence expressed in a
particular semantic language ({\em S-L}). The natural language
characteristics are generally unknown, while the semantic language
designed to cover the semantic of the application is completely
known and described by a formal grammar. The second step consists
in the translation of the sentence in {\em S-L} into computer
language code {\em C-L} for performing the requested action.
This second module can be generally (but not necessarily)  designed to 
cover all the possible sentences in {\em S-L}, since both {\em S-L}
and {\em C-L} are known. However, the boundary between the first and
the second module is quite arbitrary. In~\cite{kuhn-93}, for instance,
an automatic system is designed for translating ATIS English sentences
directly into the SQL query, and in~\cite{gorin-91} there is an
example of a system that goes from an English sentence to the requested action 
without any intermediate representation of the meaning. However, the closer
we move the definition of {\em S-L} to {\em N-L}, the more complicate
becomes the design of the {\em action transducer}, reaching in the
limit the complexity of a complete understanding system. Conversely,
when we move the definition of {\em S-L} closer to {\em C-L}, we may
find that learning the parameters of the {\em semantic translator}
becomes quite a difficult problem when the application entails
a rather complex semantics.
The subject of this paper deal with
the investigation of the possibility of automatizing the design
of the first block (i.e. the {\em semantic translator})
starting from a set of examples. The semantic language chosen
for the experiments reported in this paper is very simple and
consists of sequences of keyword/value pairs (or {\em tokens}). 
There is no
syntactic structure in the semantic language we use. Two sentences for which
the difference in the semantic representation is only in the
order of the  tokens are considered equivalent. In this way we cover a
good percentage of sentences in the domain, but still there are
sentences that would require a structured semantic language. For
instance the two following sentences are indistinguishable when
represented by our semantic language, and obviously they have
a different meaning.
\begin{quote}
{\it IS THE EARLIEST FLIGHT GOING TO BOSTON ON A SEVEN FOUR SEVEN}
\end{quote}

\begin{quote}
{\it IS THE EARLIEST FLIGHT ON A SEVEN FOUR SEVEN GOING TO BOSTON}
\end{quote}

The representation of this kind of sentences requires a more sophisticated
semantic language that allows the use of bracketing for delimiting
the scope of modifiers. 

Although the system we propose uses a very simple intermediate semantic
representation, we showed that it can successfully handle most of
the sentences in a database query application like the ATIS task.
When this simple representation is used and when the  problem of 
{\em semantic translation} is formalized as a 
communication problem, a MAP criterion can be established for 
decoding the {\em units of meaning} from text or speech. The
resulting decoder can then be integrated with other modules for
building a speech/text understanding system. 

An understanding system based on a learning paradigm, like
the one proposed in this paper, can evolve according to different
dimensions of the problem. One dimension goes with the increase in
complexity of the semantic language {\em S-L}. Rather than using
a sequential representation on could think of a tree representation
of the meaning. However, this
poses additional problems both in the training
and decoding stage, and requires the use of algorithms designed for
context-free grammars, like for instance the {\em inside-outside}
algorithm~\cite{baker-79}\cite{pereira-92} that have a higher complexity
that those explained in this paper. 
Another dimension of the problem goes toward a
complete automatization of the system, also for those modules
that, at the moment, require  a manual compilation of some
of the knowledge sources. One of these modules is the {\em template generator}.
Both~\cite{giachin-92} and~\cite{kuhn-93} report examples of systems where
the decision about the actual values of the conceptual entities 
(or an equivalent information) is drawn on the basis of knowledge
acquired automatically from the examples in the training corpus.
The kind of annotation required for the training corpus is also
another dimension along with the research on learning to understand
language should move. A strategy for learning the understanding function
of a natural language system becomes really effective and competitive
to the current non-learning methods when the amount of labor required
for annotating the sentences in a training corpus is comparable or
inferior to the amount of work required for writing a grammar in
a traditional system. This requires the development of a learning
system the does not require any other information than the representation
of the meaning associated to each sentence (e.g. it does not require an 
initial segmentation into
conceptual units, like in CHRONUS, for bootstrapping the conceptual
models). Moreover, the representation of the meaning should be made
using a {\em pseudo-natural} language, for making easier and less
time consuming the work of the annotators. An example of this
kind of annotation was introduced in section~\ref{sect:pseudoE} with
the pseudo-English {\em win} rephrasing. This suggests a possible
evolution of the learning strategy for understanding systems toward
a system starting with the limited amount of knowledge required for
understanding a small subset of the whole language (e.g. the {\em win}
language). Then the system can evolve to understanding larger subsets of the
language using the language already acquired for rephrasing new
and more complex examples. But, of course, the science of learning
to understand is still in its infancy, and many more basic problems
must be solved before it becomes an established solution to the design of
a language interface.

\end{document}